\documentclass[conference]{IEEEtran}
\IEEEoverridecommandlockouts
\usepackage{cite}
\usepackage{booktabs}
\usepackage{amsmath,amssymb,amsfonts}
\usepackage{caption}
\usepackage{graphicx}
\usepackage{textcomp}
\usepackage{xcolor}
\usepackage{multirow}
\usepackage{subcaption}
\usepackage{algorithm}
\usepackage[noend]{algpseudocode}
\usepackage{hyperref}       
\usepackage{url}            
\def\BibTeX{{\rm B\kern-.05em{\sc i\kern-.025em b}\kern-.08em
    T\kern-.1667em\lower.7ex\hbox{E}\kern-.125emX}}

\begin{document}


\title{On Large Visual Language Models for Medical Imaging Analysis: An Empirical Study}

\author{
\IEEEauthorblockN{Minh-Hao Van, Prateek Verma, Xintao Wu}
\IEEEauthorblockA{
\textit{University of Arkansas}\\
Fayetteville, AR, USA \\
\{haovan, prateek, xintaowu\}@uark.edu}
}

\maketitle

\begin{abstract}
Recently, large language models (LLMs) have taken the spotlight in natural language processing. Further, integrating LLMs with vision enables the users to explore emergent abilities with multimodal data. Visual language models (VLMs), such as LLaVA, Flamingo, or CLIP, have demonstrated impressive performance on various visio-linguistic tasks. Consequently, there are enormous applications of large models that could be potentially used in the biomedical imaging field. Along that direction, there is a lack of related work to show the ability of large models to diagnose the diseases. In this work, we study the zero-shot and few-shot robustness of VLMs on the medical imaging analysis tasks. Our comprehensive experiments demonstrate the effectiveness of VLMs in analyzing biomedical images such as brain MRIs, microscopic images of blood cells, and chest X-rays.
\end{abstract}

\begin{IEEEkeywords}
visual language model, zero-shot learning, medical imaging analysis
\end{IEEEkeywords}

\section{Introduction}
In healthcare, medical images such as magnetic resonance images, X-rays, or microscopy images are commonly used to give visual information about the human body, which a specialist can leverage to analyze the disease and plan appropriate treatments. However, analyzing medical images is a non trivial task and requires a lot of expertise. Further, due to uncertainties inside the scanned image such as poor resolution or noise, misdetection of diseases unfortunately leads to wrong or delayed treatments, which affects the survival of the patients. Recently, deep learning methods have been proposed to minimize the misclassification in image processing systems using CNN-based or transformer-based networks \cite{sahlol2020efficient,liu2021automatic}.

Visual language models, which are trained on a huge database of human knowledge, have shown incredible performance in delivering insightful clues to doctors or healthcare specialists. By leveraging the knowledge from millions to billions of training examples, VLMs can assist in detecting small abnormalities inside low-resolution images, which are hard to observe by the naked eye. Moreover, pretrained VLMs such as CLIP\cite{radford2021learning}, Flamingo\cite{alayrac2022flamingo}, LLaVA\cite{liu2023visual} or ChatGPT-4\cite{openai2023gpt4} can enable the emergent abilities on unseen tasks for which they are not specifically trained. For example, we will show that LLaVA pretrained on multimodal image-instruction pairs collected from general sources can achieve impressive performance in analyzing medical images like brain MRI or chest X-ray. Models pretrained on high-quality task-specific datasets such as BiomedCLIP\cite{zhang2023large} are released to provide more useful applications to domain-specific users.

In this work, we offer an empirical evaluation to show the performance of state-of-the-art pretrained VLMs in analyzing medical imaging via zero-shot and few-shot prompting (without the need for model retraining or fine-tuning). To the best of our knowledge, this work is one of the very first attempts to conduct a comprehensive evaluation to study the ability of VLMs on different medical imaging datasets. Prior to our work, Yan et al \cite{yan2023multimodal} experimentally studied multimodal ChatGPT for medical applications. However, the authors mainly explored ChatGPT-4V with visual question-answering (VQA) tasks. We instead evaluate the classification task of five different VLMs on three different image types, i.e., MRI, microscopy, and X-ray. We also show the reasoning ability of some visual language chat models like LLaVA or ChatGPT4 on how they can come up with the answer. Finally, we discuss the limitations of VLMs when working on medical data. While there is still a lack of exploratory works on this problem, we hope that our work will provide interesting and insightful findings to the multi-disciplinary research community.


\section{Related Work}
\label{sec:related-work}
In the past few years, decent efforts have been made towards assessing the prowess of large language models in the area of medicine. Med-PaLM 2 \cite{singhal2023towards} exhibited 86.5\% accuracy respectively on MedQA, a USMLE style dataset. LLM chatbots sometimes respond with impressive answers, demonstrating that they are able to dig relevant clinical information, but are still far from being trustworthy and safe \cite{thirunavukarasu2023large}. A few studies have recently used VLMs in the analysis of medical images. Wang et al \cite{wang2023chatcad} integrated existing LLMs with computer-aided diagnostic (CAD) networks for diagnostic, segmentation, and reporting tasks. 

In terms of models that haven't been specifically trained or fine-tuned on medical image data, most recently, Yan et al evaluated the performance of the vision-capable GPT, the ChatGPT-4V, on simple medical VQA tasks and concluded that ChatGPT-4V in its current state was not fit for real-world diagnosis \cite{yan2023multimodal}. Specific to the datasets that we have chosen for this study, a few diagnostic tasks have been attempted using LLMs or VLMs. For example, Ahmed et al \cite{allah2023edge} utilized the BTD dataset\cite{Cheng2017} and developed an Edge U-Net model for the precise location of tumors. 
Several works that have used the ALL-IDB2 dataset \cite{labati2011all} for classification tasks include hybrid CNN \cite{sahlol2020efficient} or transfer learning with SVMs \cite{vogado2018leukemia}. Recently, Liu et al \cite{liu2023medical} demonstrated the progression of COVID-19 disease using a multimodal (vision and language) large language model. 

\section{Visual Language Models}
\label{sec:vlms}

\noindent\textbf{OpenCLIP.} This is an open-source version of CLIP \cite{radford2021learning}, proposed to reimplement and reproduce the proposed method. CLIP (Contrastive Language-Image Pre-Training) is a neural network connecting the image and text. The target is to learn the relationship between the images and their textual description (captions or labels). The model includes two main components: an image encoder and a text encoder. The image encoder, which is usually based on Vision-Transformer\cite{dosovitskiy2020image} or ConvNext\cite{liu2022convnet} architectures, produces the visual features of the input image in the embedding space. The text encoder, which is based on Transformer\cite{vaswani2017attention} architecture, aims to produce the embedding of input text. The method predicts the likelihood of classes by calculating the similarity score between the image embedding and all text embeddings. 

\begin{figure}[ht]
    \centering
    \includegraphics[width=.30\textwidth]{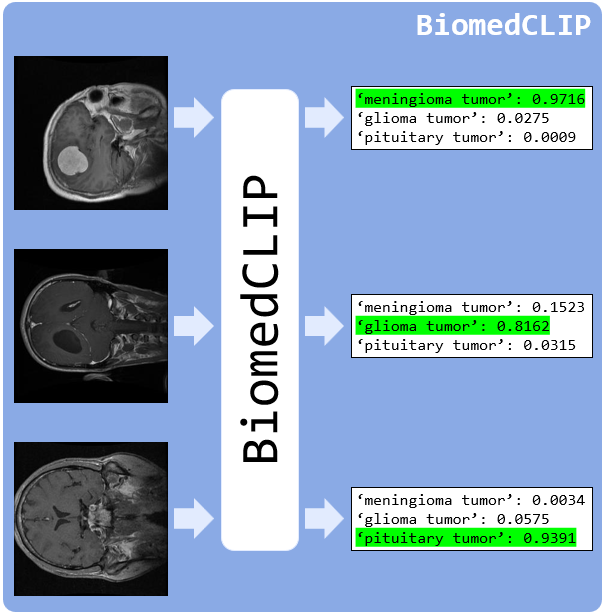}
    \caption{An example of BiomedCLIP in predicting brain tumor type from MRIs. Green highlighted text indicates correct prediction by the method.}
    \label{fig:biomed_exp}
\end{figure}

\noindent\textbf{BiomedCLIP.} Standard versions of CLIP are originally trained on general image classification tasks, which still have a gap in performance on domain-specific data such as biomedical images. To alleviate the gap, \cite{zhang2023large} proposed BiomedCLIP as an adaptation of CLIP to the biomedical field, which is trained on a large-scale image-text dataset. In Figure \ref{fig:biomed_exp}, we show some examples of how BiomedCLIP classifies tumor types in brain MRIs. In the figure, we test three input images that belong to three different types of tumor (meningioma, glioma, and pituitary). Similarly, we create three input texts crafted according to the expected classes: ``meningioma tumor", ``glioma tumor" or ``pituitary tumor". For each image, BiomedCLIP outputs probabilities of image-text matching pairs. In the examples, the first MRI has 97.16\%, 2.75\%, and 0.09\% probabilities belonging to meningioma, glioma, and pituitary classes, respectively.

\begin{figure}[ht]
    \centering
    \includegraphics[width=.30\textwidth]{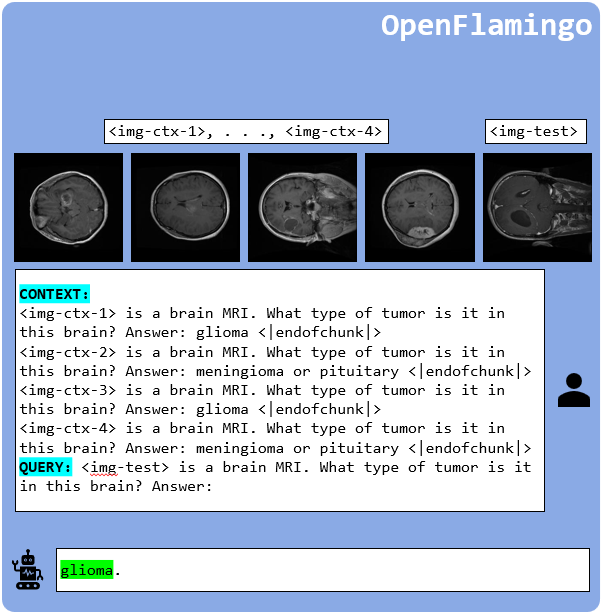}
    \caption{An example of OpenFlamingo in predicting brain tumor type from MRIs. Green highlighted text indicates correct prediction by the method.}
    \label{fig:of_exp}
\end{figure}

\noindent\textbf{OpenFlamingo.} This is an attempt to replicate Flamingo\cite{alayrac2022flamingo} and make it open-source. Flamingo executes vision-language tasks effectively using pretrained visual encoders and language models. To connect those modules, the Perceiver Resampler is trained to project extensive embedding features from the encoder to visual tokens. Then, both visual and textual tokens are combined and sent to the pretrained language model. The method further modifies the pretrain LLM by adding gated cross-attention layers. Therefore, the cross-attention scheme allows the model to process multiple image-text inputs, enabling the ability to perform in-context learning. In Figure \ref{fig:of_exp}, we demonstrate the usage of OpenFlamingo with a brain tumor classification task. To do few-shot learning, we use four images from the training data to build the context and test the performance on a different test image that has not been seen by the model. The model works as a text generation where it will finish the query provided by the user. In the figure, brain MRIs and their corresponding text descriptions are sent to the model as demonstration examples. We expect the model to learn how to classify tumor types inside the images and predict the tumor type of the test image. In the example, the model can successfully provide the correct answer.

\begin{figure}[ht]
    \centering
    \includegraphics[width=.30\textwidth]{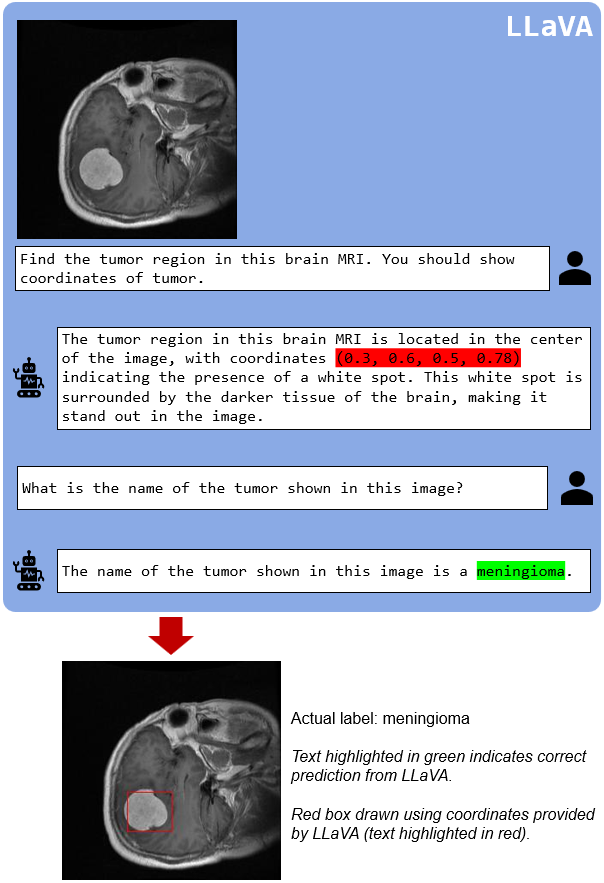}
    \caption{An example of LLaVA in predicting brain tumor type from MRIs.}
    \label{fig:llava_exp}
\end{figure}

\noindent\textbf{LLaVA.} Different from the previous method, LLaVA\cite{liu2023visual} is trained as a visual language assistant using multimodal instruction-following data. Specifically, LLaVA works as a chat assistant that provides human-like conversation to users. With the integration of a vision encoder, LLaVA can understand the image input and generate accurate answers to multiple visio-linguistic tasks. To do so, this VLM combines pretrained CLIP and Llama-2\cite{touvron2023llama} as the visual encoder and language model, and uses a projection layer to map image features to language tokens. Then, both visual and textual tokens are used as inputs to the language model. To demonstrate the ability of LLaVA, we show an example in Figure \ref{fig:llava_exp}. As a multimodal chat assistant, LLaVA can serve multiple query-answer pairs. We first test the image understanding ability of LLaVA by asking it to find the tumor region inside the image. While the model was not specifically trained on the brain MRI dataset, it can give an impressive answer with precise coordinates of the tumor area. The answer additionally comes with an explanation that the model can know the abnormal area by pointing out dark tissue and white spot. Based on that conversation, we then request LLaVA to classify the tumor type and the answer is correct. The chain of thought\cite{wei2022chain} built upon multiple sub-questions is also a technique that can further enhance the performance of the language model. 

\begin{figure}[ht]
    \centering
    \includegraphics[width=.30\textwidth]{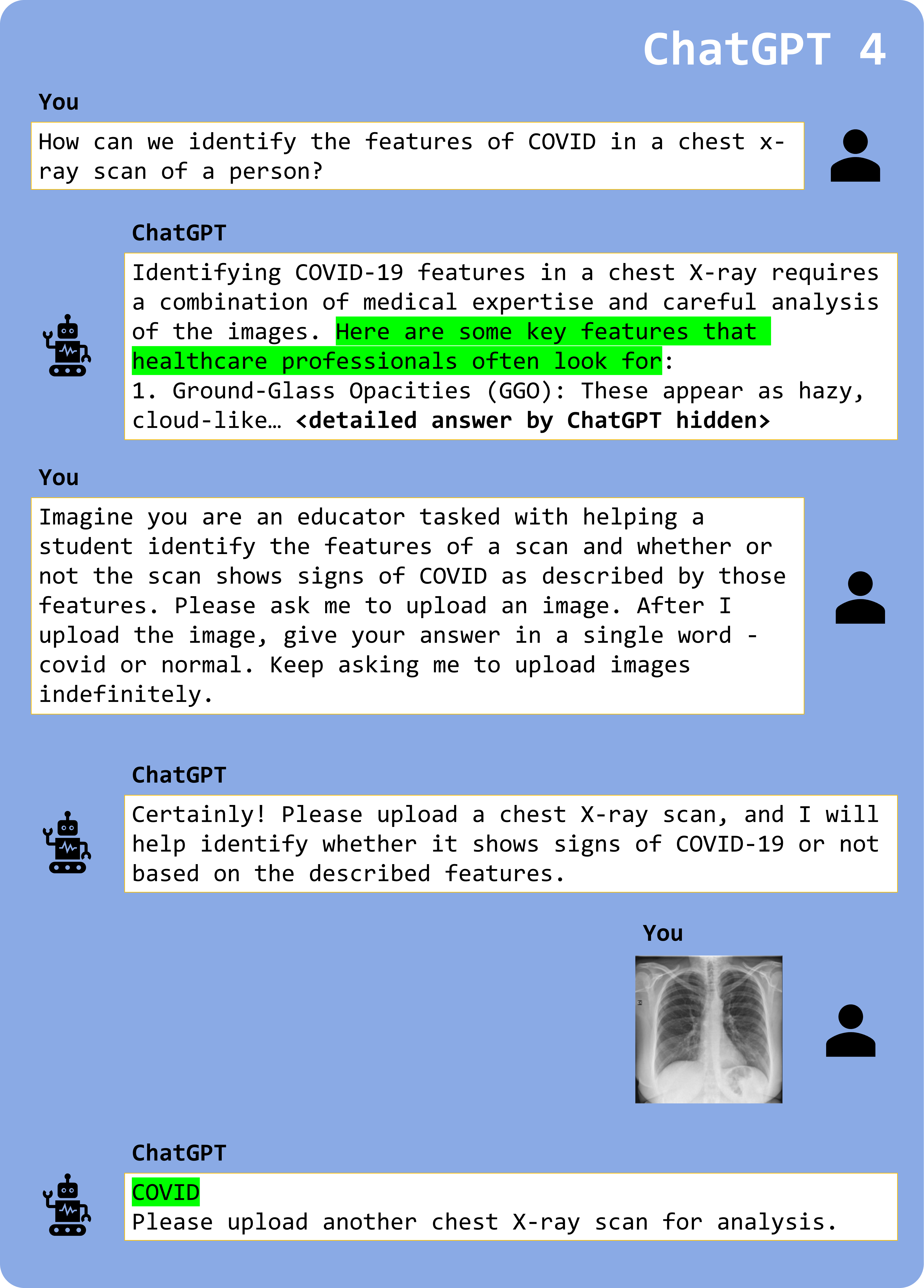}
    \caption{An example of ChatGPT-4 in predicting COVID-19 from scans of a chest X-ray.}
    \label{fig:chatgpt_exp}
\end{figure}

\noindent\textbf{ChatGPT-4.} With GPT-4, OpenAI introduced the ability to input both images and text together to produce text output. GPT-4 is a pre-trained Transformer-based model that outperformed GPT-3.5 on most academic and professional exams and other state-of-the-art models on various academic benchmarks and hallucinates less \cite{openaiGPT4TechnicalReport2023}. Figure \ref{fig:chatgpt_exp} shows an example of the ChatGPT-4 interface wherein ChatGPT was asked how to identify the symptoms of COVID in an X-ray of the chest. ChatGPT's answer shows that it was well-versed in identifying features of COVID-19 in an X-ray. Subsequently, ChatGPT is asked to provide its diagnosis on the next uploaded image in a single word, which it complied with, and answer correctly.

\section{Experiment Results}
\label{sec:experiments}
In this section, we evaluate the classification performance of five VLMs (BiomnedCLIP, OpenCLIP, OpenFlamingo, LLaVA, ChatGPT-4) and two CNN-based baselines (CNN, ResNet-18) on three medical imaging datasets (BTD, ALL-IDB2, CX-Ray). For CNN-based methods, we first train the models on the training set and then evaluate them on the test set. For VLMs, as there is no need for training or fine-tuning the models, we directly evaluate the models on a test set with zero-shot (BiomedCLIP, OpenCLIP, LLaVA, ChatGPT-4) and few-shot (OpenFlamingo) prompting. Except for OpenFlamingo, we also use the training set to build the demonstration set. Note that there is no fine-tuning with OpenFlamingo because it only leverages the demonstration set as contextual information for adaptation to the test image. We report the test accuracy (\%) on each dataset and the average accuracy of three datasets. 
The details of the three datasets are shown in Table \ref{tab:data_details}.

\subsection{Datasets}
\label{sec:dataset}

\begin{table}[ht]
    \centering
    \caption{Description of datasets.}
    \begin{tabular}{l|c|c|c}
    \toprule
         Dataset & Class & Train  & Test \\ \midrule
         \multirow{2}{*}{BTD} & Meningioma or Pituitary & 1220 & 303 \\ 
         & Glioma & 1037 & 325 \\ 
        \midrule
        \multirow{2}{*}{ALL-IDB2} & Normal & 78 & 52 \\
        & Blast & 78 & 52 \\ \midrule
        \multirow{2}{*}{CX-Ray} & Negative & 160 & 27 \\
        & Positive & 264 & 67 \\
        \bottomrule
    \end{tabular}
    \label{tab:data_details}
\end{table}

\noindent\textbf{Brain Tumor Detection (BTD)\cite{Cheng2017}.} The dataset contains 3,064 T1-weighted CE-MRI slices from 233 patients. The task is to classify the tumor types (glioma, meningioma, or pituitary) inside images. To do the binary classification task, we merge samples with ``meningioma" and ``pituitary" labels into ``meningioma or pituitary" labels, and samples with `gliomas" labels remain the same.

\noindent\textbf{Acute Lymphoblastic Leukemia Image Database (ALL-IDB)\cite{labati2011all}.} This microscopic image dataset focuses on a blood pathology, called Acute Lymphoblastic Leukima. ALL-IDB2 (260 images) includes a cropped area of interest of particular cells for testing the classification of normal or blast cells.

\noindent\textbf{COVID Chest X-ray (CX-ray)\cite{cohen2020covid}.} The target is to build a publicly open dataset of chest X-ray and CT images for analyzing COVID-19, and other viral or bacterial pneumonias. In our evaluation, the task is to predict whether the patient is positive or negative for COVID-19. 
\subsection{Results}
\label{sec:results}
\begin{table*}[t]
    \centering
    \caption{Test accuracy (\%) of VLMs and baselines on different medical imaging datasets. Boldfaced value indicates the best accuracy among VLMs on each dataset}
    \begin{tabular}{l|ccc|cccc}
        \toprule
        \multirow{2}{*}{Type} & \multirow{2}{*}{Model} & \multirow{2}{*}{Architecture} & \multirow{2}{*}{Shots} & \multicolumn{3}{c}{Datasets} & \multirow{2}{*}{Average} \\
        & & & & BTD & ALL-IDB2 & CX-ray &  \\ \midrule
        \multirow{2}{*}{CNN-based} & CNN\cite{lecun1989backpropagation} & 5$\times$CNN+1$\times$FC & 0 & 92.52 & 92.31 & 92.55 & 92.46\\
        & ResNet-18\cite{he2016deep} & 17$\times$CNN+1$\times$FC & 0 & 91.05 & 94.23 & 93.62 & 92.97 \\ \midrule
        \multirow{2}{*}{VLM} & BiomedCLIP\cite{zhang2023large} & ViT-B/16 & 0 & \textbf{79.14} & 76.92 & 58.51 & \textbf{71.52}\\
        & OpenCLIP\cite{ilharco_gabriel_2021_5143773} & ViT-G/14 & 0 & 57.96 & 69.23 & 63.83 & 63.67\\
        & LLaVA\cite{liu2023visual} & ViT-L/14 \& LLaMA-2-7B\cite{touvron2023llama} & 0 & 54.46 & 66.72 & 65.18 & 62.12\\
        & ChatGPT-4 \cite{openai2023gpt4} & GPT-4 & 0 & 51.61 & \textbf{84.85} & 61.82 & 66.09 \\
        & OpenFlamingo\cite{awadalla2023openflamingo} & ViT-L/14 \& INCITE-3B & 0 & 49.68 & 55.77 & 62.77 & 56.07\\
        &  &  & 2 & 50.48 & 67.31 & 68.44 & 62.08\\
        &  &  & 4 & 51.75 & 50.00 & \textbf{71.28} & 57.68\\
        \bottomrule
    \end{tabular}
    \label{tab:performance}
\end{table*}

\subsubsection{Overall Performance} In Table \ref{tab:performance}, we show the performance on VLMs and baselines on benchmarked datasets. CNN-based methods achieve the best performance on all datasets, which is understandable as CNN-based methods are pre-trained with the training set while LLMs are not. However, pretrained VLMs still have impressive performance, at no cost for the training stage - using only pretrained models. Further, with CNN-based methods, users need to train three separate model instances for three different datasets while only one VLM instance is needed for all datasets. Among VLMs, BiomedCLIP, ChatGPT, and OpenFlamingo are the best performers with BTD, ALL-IDB2, and CX-Ray, respectively. Overall, BiomedCLIP achieves the best performance averaging on three datasets.

\subsubsection{Prompting VLMs to Medical Imaging Classifcation}
VLMs with vision input capabilities are used for evaluating their zero-shot performance. VLMs can be inconsistent as they have billions of parameters and rely on the statistical relationship of words to predict the next word based on the provided prompt, which may raise the randomness in the responses. It is commonly known that prompting a VLM to perform well on unseen tasks is not trivial. Prompt engineering is a strategic task to optimize the input prompt such that the VLM can give good and meaningful answers. In this section, we discuss how to build robust prompts for adapting VLMs to medical imaging analysis. 

As aforementioned in Section \ref{sec:vlms}, \textbf{OpenCLIP} and \textbf{BiomedCLIP} predict the class by calculating similarity-based scores between the input image and class descriptions. A typical template for CLIP-based methods is \textit{``a photo of [class]"}, which however does not lead to the best performance for these methods. Therefore, we build specific prompts for each dataset. With BTD, \textit{``glioma tumor"} and \textit{``meningioma or pituitary tumor"} are the best text descriptions of the classes. Similarly, we craft the prompts for ALL-IDB2 as \textit{``normal cell"} and \textit{``Acute Lymphoblastic Leukemia lymphoblast"}. With CX-Ray, more detailed prompts are required to achieve a good performance, which is \textit{``The chest X-Ray does not show any sign of COVID"} and \textit{``The chest X-Ray shows a sign of COVID"}. Although BiomedCLIP is not among the top performers with CX-Ray, it outperforms all other VLMS on average. This result is not surprising as BiomedCLIP is the only VLM that was trained on a domain-specific dataset for biomedical imaging.

With \textbf{OpenFlamingo}, we try both zero-shot and few-shot prompting techniques. Note that this is the only VLM that is able to execute few-shot prompting - receiving multiple input images - among methods used in our study. Similar to the example in Figure \ref{fig:of_exp}, for the BTD dataset, we prompt the model with the template \textit{``\textless image\textgreater\ is a brain MRI. What type of tumor is it in this brain? Answer: [class]"}. For CX-Ray, we format the demonstration as \textit{``\textless image\textgreater\  This is a [class] chest X-ray"}, in which the class is either normal or COVID-19. Similarly, the demonstration of ALL-IDB2 is formatted as \textit{``\textless image\textgreater\  is this normal or leukemia cell? Output: [class]"}. For few-shot prompts, we randomly choose the demonstration images from the training set. Looking at the results of zero-shot and few-shot prompting techniques, one can observe that the demonstration from few-shot prompts helps improve accuracy in most cases. Except for ALL-IDB2, an increasing trend in accuracy is clear from the results. Subsequently, determining the appropriate number of shots and selection strategy of demonstration are worthy of further exploration.

\textbf{LLaVA}, a visual chat assistant, can generate human-like responses. For the single-step query, we craft a single prompt to describe the task, data, and expected response. For example, with the BTD dataset, one useful prompt is \textit{``You are a radiologist. You are presented with a brain MRI scan of a patient with a tumor. Please diagnose whether the patient shows signs of a glioma or some other type of tumor. Please provide your answer as either in a single word - `glioma' or `other'."}. In the prompt, we first let LLaVA know that it would pretend to be a radiologist. The purpose of this instruction is to reduce the hallucination of LLaVA and tell it to focus on radiological image diagnosis. Then, LLaVA is asked to analyze the input image and produce the answer in the format (`glioma' or `other'). Different from previous VLMs that are prompted with single-step query, we further try multi-step prompt with LLaVA. The example shown in Figure \ref{fig:llava_exp} demonstrates the ability of LLaVA to handle multiple requests in one inference. The previous question-answer pairs can serve as the context for the next question as a chain of thought, which can significantly enhance the performance.  

With \textbf{ChatGPT}, a multitude of prompts were tried to narrow down on a prompt style that could elicit a single-word response. A multi-step prompting technique, shown in Figure \ref{fig:chatgpt_exp}, was found to consistently produce a single-word response from the model, and the best accuracy in most cases. ChatGPT is first asked about how it would go about analyzing the medical image for the chosen classes. Thereafter, we put ChatGPT in the role of an educator and ask it to perform the classification. After uploading the image, ChatGPT responds with the class prediction. In the specific case of All-IDB2, ChatGPT performs better with the single-step prompting, wherein it was asked to identify whether the image shows signs of a blast cell or not, resulting in the reported 84.85\% accuracy. An accuracy of 75.76\% was obtained with the multi-step prompting technique.
\section{Conclusion}
\label{sec:conclusion}
\noindent\textbf{Limitation of VLMs.}
Although visual language models can provide interesting and useful applications to users, there is a gap in a fully automated AI system. Uncertainty in VLMs is the quality of data, the safety of their response, and the potential security or privacy exploitation. While VLMs are helpful in language tasks like grammar check, question answering, or information retrieval, they are not ready to be used as a replacement for humans in domain-specific applications such as analyzing and predicting disease from medical images. The data collection and data usage are other limitations for VLMs. The performance of AI models relies on the data because of 
the quality of output strongly depends on the quality of input. More and more, the use of training data has raised privacy concerns recently as they may unintentionally leak private information learned from data. 

\noindent\textbf{Discussion.} We study the use of VLMs in classifying medical images. Although VLMs can not outperform classic deep learning models like CNN or ResNet on benchmarked datasets, it is worth noting that they can serve as chat assistants to provide pre-diagnosis before making decisions. We hope that our study can provide helpful insights to the community. In the future, we will explore more tasks such as segmentation using state-of-the-art VLMs that are specifically trained for biomedical imaging analysis. 

\section*{Acknowledgements}
This work was supported  in part by National Science Foundation under awards 1946391, the National Institute of General Medical Sciences of National Institutes of Health under award P20GM139768, and the Arkansas Integrative Metabolic Research Center at University of Arkansas.

\end{document}